\documentclass[runningheads]{llncs}
\usepackage[T1]{fontenc}
\usepackage{cite}
\usepackage{amsmath}

\usepackage{graphicx}
\begin{document}
\title{Actions and Objects Pathways for Domain Adaptation in Video Question Answering}
%
%
\author{Safaa Abdullahi Moallim Mohamud\inst{1}\orcidID{0000-0002-2380-9772} \and
Ho-Young Jung\inst{2}\orcidID{0000-0003-0398-831X}}
%
%
\institute{Center for ICT and Automotive Convergence, Kyungpook National University, Daegu, 41566, Republic of Korea \and Graduate School of Artificial Intelligence, Kyungpook National University, Daegu, 41566, Republic of Korea}
\maketitle              
\begin{abstract}
In this paper, we introduce the Actions and Objects Pathways (AOPath) for out-of-domain generalization in video question answering tasks. AOPath leverages features from a large pretrained model to enhance generalizability without the need for explicit training on the unseen domains. Inspired by human brain, AOPath dissociates the pretrained features into action and object features, and subsequently processes them through separate reasoning pathways. It utilizes a novel module which converts out-of-domain features into domain-agnostic features without introducing any trainable weights. We validate the proposed approach on the TVQA dataset, which is partitioned into multiple subsets based on genre to facilitate the assessment of generalizability. The proposed approach demonstrates 5\% and 4\% superior performance over conventional classifiers on out-of-domain and in-domain datasets, respectively. It also outperforms prior methods that involve training millions of parameters, whereas the proposed approach trains very few parameters. 


\end{abstract}

\section{Introduction}

Video Question Answering (Video QA) represents a pivotal task in deep learning, allowing systems to comprehend visual content in response to natural language queries \cite{Gao_2023_CVPR}. 
The adoption of large pretrained models is increasingly becoming the norm in VideoQA due to their capacity to adapt to various downstream tasks \cite{NEURIPS2021_c6d4eb15}. 
However, the large size of pretrained models and the significant computational resources they demand pose challenges for fine-tuning with limited resources \cite{Bain_2021_ICCV}. Without access to substantial computational resources, leveraging the capabilities of large pretrained models becomes challenging. Moreover, these large pretrained models does not exhibit excellent adaptability to other domains without the fine-tuning of the entire framework \cite{NEURIPS2022_00d1f03b}. This fine-tuning process is computationally intensive, often requiring significant resources and time. As illustrated in Table \ref{tab:intro_zero_shot}, FrozenBiLM, which is a robust pretrained framework trained on over 10 million video-text pairs, does not exhibit comparable performance on the TVQA dataset in a zero-shot manner, unlike the fine-tuned framework \cite{NEURIPS2022_00d1f03b}. Similarly, Merlot Reserve, pretrained on more than 20 million YouTube videos, fails to deliver comparable results when the framework weights are frozen and only the classifier is trained, compared to fine-tuning the entire framework \cite{Zellers_2022_CVPR}. Nevertheless, these large models inherently retain rich features acquired from substantial amounts of training data \cite{Chen_2022_CVPR}. Therefore, it is imperative to seek computationally inexpensive approaches to harness the potential of pretrained features without the need for fine-tuning the entire large model. 
\begin{table}
\begin{center}
\caption{Validation experiments of large pretrained frameworks demonstrating accuracy discrepancies between fine-tuned and unfine-tuned models}\label{tab:intro_zero_shot}
\begin{tabular}{cccc}
\hline
Framework &  Accuracy & Framework &  Accuracy \\
\hline
FrozenBiLM &  59 &Reserve + Trained ATClassifier &  64\\
Finetuned FrozenBiLM & \textbf{82} &Finetuned Reserve + Trained ATClassifier &  \textbf{83}\\
\hline
\end{tabular}
\end{center}
\end{table}

In this view, we introduce the Actions and Objects Pathways (AOPath) task-specific classifier for robust VideoQA tasks, with the aim of harnessing the rich features of large pretrained models \cite{Zellers_2022_CVPR}. The AOPath task-specific classifier incorporates concepts of human biological cognition, specifically action and object gates \cite{wurm2022two}, into its design to provide a solution to computationally inexpensive generalizability. In the human brain, there are two pathways to process sensory data: the object pathway and the action pathway. The object pathway, primarily associated with the ventral stream, processes visual information related to objects and shapes. On the other hand, the action pathway, mainly associated with the dorsal stream, focuses on processing information related to motion and actions \cite{lingnau2015lateral}. The integration of these pathways is facilitated by mechanisms like action and object gates which enable the brain to integrate the recognition of actions and objects in human perceptual experience.

\begin{figure}[tb]
  \centering
  \includegraphics[height=7cm]{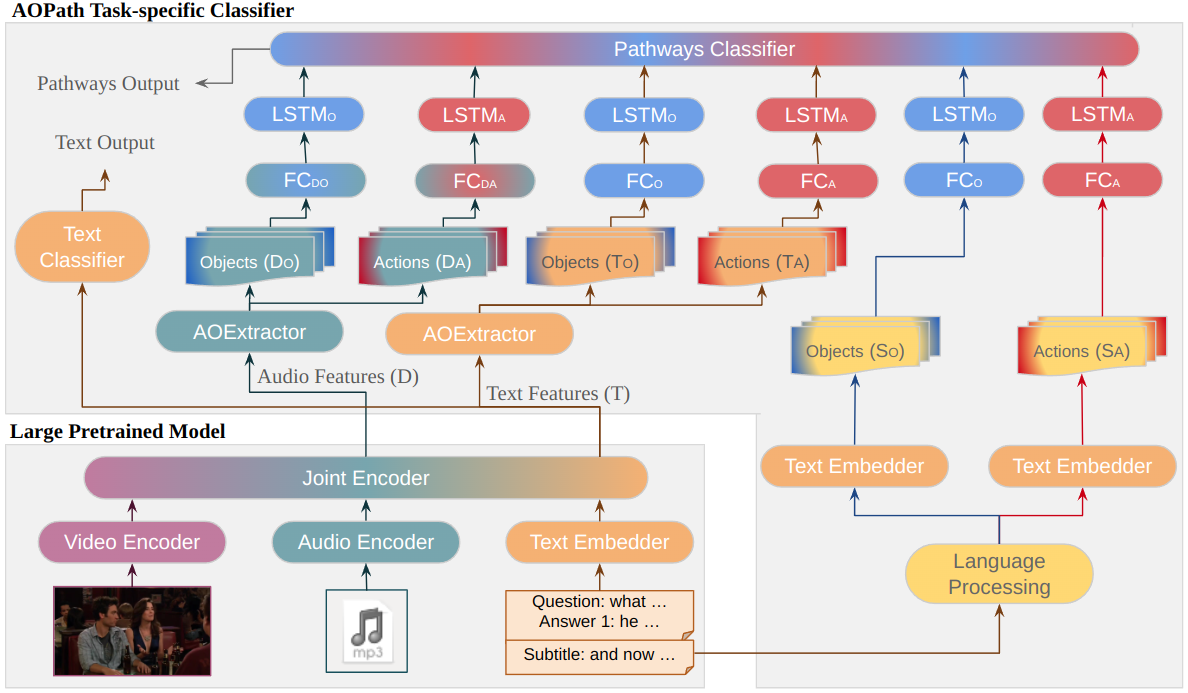}
  \caption{The proposed AOPath task-specific classifier. 
  }
  \label{fig:framework}
\end{figure}

The proposed AOPath is a novel approach that generalizes to out-of-domain datasets, thereby reducing the need for additional fine-tuning tailored to unseen domains. As illustrated in Fig. \ref{fig:framework}, it employs a unique approach of decomposing pretrained features into distinct, rich action and object features, and subsequently processes them in separate reasoning pathways. It employs the AOExtractor module, which has no trainable weights, to extract domain-agnostic features by selecting the most similar action and object labels from predefined dictionaries. Moreover, it similarly extracts action and object words from the associating subtitle to create its distinct reasoning pathways. Lastly, in the pathways classifier, it obtains the similarities among features in each pathway to infer the correct answer introducing very few additional trainable weights. The small version of the proposed approach, $AOPath_S$, has only 26,282 trainable weights, stemming from the FC and LSTM modules employed to process each pathway. The AOExtractor and Pathways Classifier modules have very few to no trainable weights.

We validate the proposed task-specific classifier on the TVQA dataset \cite{lei2019tvqa}, which comprises a diverse collection of video clips from 6 popular television shows, accompanied by human-written question-answer pairs, providing a rich source for training and evaluating robust VideoQA systems. In our experimental setup, we partition the dataset into three distinct subsets based on the genre of the TV series. Subsequently, we train the proposed classifier on one genre and validate its performance on another to assess its generalizability across diverse domains and storytelling styles. The proposed approach exhibits superior performance in terms of generalizability compared to a conventional classifier, outperforming it by 5\% and 4\% in out-of-domain and in-domain datasets, respectively. The small AOPath also showcases a 3\% improvement in performance on out-of-domain datasets compared to previous methods that require training millions of parameters.

In this paper, we propose the Actions and Objects Pathways (AOPath) for robust VideoQA whose main contributions are: 
  \begin{itemize}

  \item Inspired by human brain, AOPath leverages pretrained features, dissociates them into action and object features, and then processes them through two distinct pathways: the actions pathway and the objects pathway.
  \item AOPath employs the innovative AOExtractor module, which dissociates and transforms out-of-domain features into domain-agnostic features without introducing any trainable weights.
  \item We validate the proposed approach, AOPath, on out-of-domain genre subsets of the TVQA validation dataset. It demonstrates superior performance compared to previous methods, achieving this with significantly fewer parameters.
\end{itemize}

\section{Related Work}


VideoQA frameworks initially encoded each modality independently \cite{yu2024self, Huang_2023_CVPR}, then subsequently merged them using cross-modal fusion modules \cite{NEURIPS2022_00d1f03b, yang2022learning}. Emerging frameworks integrated audio encoders in addition into their processing pipelines \cite{Zellers_2022_CVPR, Li_2022_CVPR_2}. Moreover, a common strategy in VideoQA models involved breaking down modality features into distinct representations. For example, video features are further partitioned into separate appearance and motion features \cite{ Xiao_Yao_Liu_Li_Ji_Chua_2022, Park_2021_CVPR}. Various other techniques divided the visual features into different components, such as complementary and casual elements \cite{Li_2022_CVPR}, long view and short view representations \cite{Ye_2023_ICCV}, and multi-scale representations \cite{10018408, peng2022multilevel}. Instead of further dividing the features, some models incorporated specialized modules to handle the different attributes in the input, such as the spatial and temporal dimensions \cite{Gao_2023_CVPR, Ye_2023_ICCV, maaz2023videochatgpt}. We argue that exploring diverse feature segmentation methods could enhance VideoQA models' robustness, leading to better performance in handling out-of-domain data. In this study, we further segment the input features into action and object features.

Transfer learning represents a strategy that has been explored in domains like visual question answering (VQA) and VideoQA to tackle the challenge of generalizability \cite{xu2019openended, Chao_2018_CVPR, wu2021transferring}. Other strategies involved subjecting the model to different variations of the training dataset to enhance its ability to generalize to out-of-distribution scenarios \cite{gokhale2020mutant}. However, our objective in this study is to achieve robustness and generalization to an out-of-domain dataset without explicitly retraining the framework on that specific domain or modifying the training dataset.


\section{Proposed Approach}

The proposed integrated framework takes as inputs a video ($v$), a subtitle ($s$), an audio ($d$), a question ($q$), and answer choices ($c_n$) to infer the correct answer. 
However, the proposed classifier leverages features obtained from a large pretrained model to encode the inputs into audio ($D$) and text ($T$) features. The audio features ($D$) encode information mainly from the audio, video, question, and answer choices, while the text features ($T$) encode information mainly from the subtitle, video, question, and answer choices. Moreover, it leverages the subtitle features as an input in two ways: first, by passing the subtitle features through the large pretrained model and integrating them with text ($T$) features; and second, by using the raw subtitles from the dataset as an independent input to extract features unrelated to the question or answer, as presented in Eq. (\ref{eq:framework2}).



\begin{equation}\label{eq:framework2}
\hat{c_n} = \mathrm{argmax} \; \theta_{AOPath}(c_n|D,T, s) 
\end{equation}

\subsection{Extraction of Action and Object Features}

\begin{figure}[tb]
  \centering
  \includegraphics[height=4.3cm]{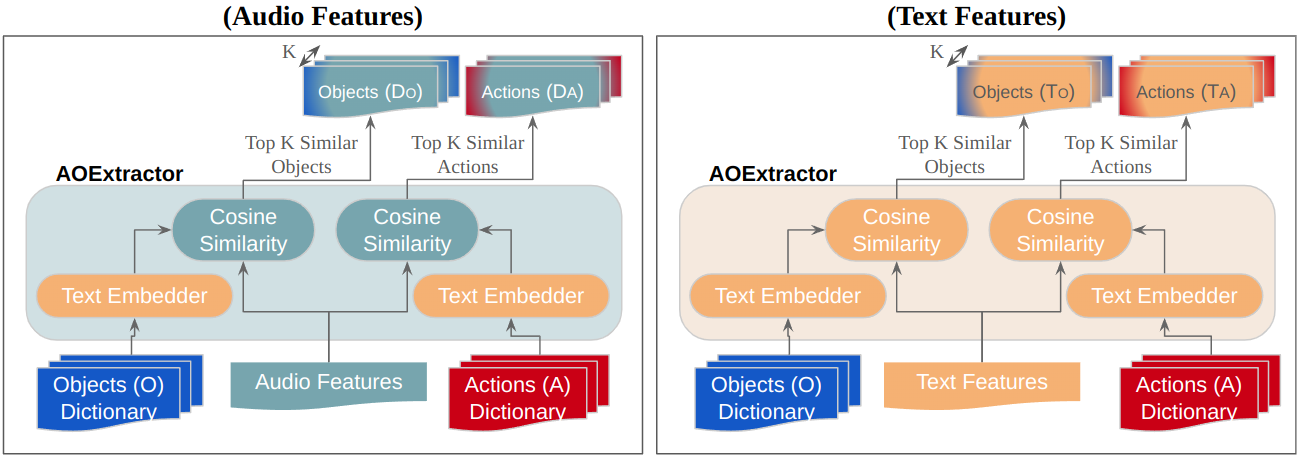}
  \caption{The AOExtractor module for audio and text features. 
  }
  \label{fig:AOExtractor}
\end{figure}

As shown in Fig. \ref{fig:framework}, pretrained features undergo processing through the Actions and Objects Extractor module (AOExtractor). The AOExtractor module extracts distinct action and object features to establish separate reasoning pathways for each feature. For each modality, the extraction of action features is achieved by computing the cosine similarity of the modality features and the embeddings of a predefined action labels, as shown in Fig. \ref{fig:AOExtractor}. Cosine similarity is employed due to its simplicity and efficiency. Unlike other similarity metrics, it measures the cosine of the angle between two vectors and is widely used in text analysis. Additionally, it is a straightforward and effective method that does not require additional trainable weights, making it perfectly aligned with the objective of the proposed AOPath. The predefined action dictionary comprises 1000 action labels, which are embedded using the pretrained text embedder of the large pretrained model. Through the cosine similarity module, the AOExtractor module identifies $K$ action labels ($A$) that exhibit the highest similarities with the modality features. Similarly, it identifies $K$ object labels ($O$) from the predefined objects dictionary that demonstrate the highest similarities with the features of the modality. These identified action and object embeddings encapsulate the prominent actions, behaviors, sensory cues, objects, and elements represented in the modality features, thereby reflecting the content depicted in the video, audio, subtitle, question, and answer choices. By employing this approach, the AOExtractor module effectively extracts domain-agnostic features and establishes the actions and objects pathways.

As presented in Eq. (\ref{eq:projecting_ao_features}), the notations $D_{A}$, $D_{O}$, $T_{A}$, and $T_{O}$ 
represent the sets of the K most similar action and object embeddings to the audio and text modalities, respectively. As illustrated in Fig. \ref{fig:framework}, the outputs of the AOExtractor module are subsequently passed to linear layers to form distinct pathway features. Four fully connected layers ($FC_{DA}$, $FC_{DO}$, $FC_A$ and $FC_O$) are employed to encode both the action and object features, as presented in Eq. (\ref{eq:projecting_ao_features}). 

\begin{equation}\label{eq:projecting_ao_features}
\begin{split}
\bar{D}_{A} = FC_{DA}(D_{A}; \theta_{FC_{DA}}),\hspace{.5cm} D_{A} = \{A^k| 0 \leq k \leq K\} \\
\bar{D}_{O} = FC_{DO}(D_{O}; \theta_{FC_{DO}}), \hspace{.5cm} D_{O} = \{O^k| 0 \leq k \leq K\}\\
\bar{T}_{A} = FC_{A}(T_{A}; \theta_{FC_{A}}), \hspace{.5cm} T_{A} = \{A^k| 0 \leq k \leq K\}\\
\bar{T}_{O} = FC_{O}(T_{O}; \theta_{FC_{O}}), \hspace{.5cm} T_{O} = \{O^k| 0 \leq k \leq K\}\\
\end{split}
\end{equation}

\subsubsection{Language Processing of Subtitles} 
The proposed classifier extracts action and object features from the subtitles that are unrelated to the question or answer. It extracts part-of-speech words that have either the tag “verb” or “noun”. To tackle the challenge of generalization, it extracts only predefined verb and noun labels present in the dictionaries from the subtitle content in the language processing phase. Extracted subtitle verbs ($V$) and subtitle nouns ($N$) undergo embedding through the pretrained text embedder of the large pretrained model, as presented in Eq. (\ref{eq:subtitle_embedder}). 

\begin{equation}\label{eq:subtitle_embedder}
\begin{split}
S_{A} = TextEmbedder(V) , \hspace{.5cm}
S_{O} = TextEmbedder(N)
\end{split}
\end{equation}

Embedded subtitle verbs ($S_{A}$) and subtitle nouns ($S_{O}$) are then forwarded to linear layers ($FC_A$ and $FC_O$) to process each pathway features separately, 
as demonstrated in Eq. (\ref{eq:projecting_subtitle_ao_features}).

\begin{equation}\label{eq:projecting_subtitle_ao_features}
\begin{split}
\bar{S}_{A} = FC_{A}(S_{A}; \theta_{FC_{A}}) , \hspace{.5cm}
\bar{S}_{O} = FC_{O}(S_{O}; \theta_{FC_{O}})
\end{split}
\end{equation}

\subsubsection{Global Representations} Following the extraction of the action and object features, the next stage involves feeding these features into bidirectional Long Short-Term Memory (LSTM) modules, as depicted in Fig. \ref{fig:framework}. The rationale for utilizing LSTM modules arises from the compact AOPath feature space. Their efficient cell and hidden vectors facilitate the retention and updating of relevant information, enabling the classifier to learn effectively even from a highly constrained feature space. These bidirectional LSTMs ($LSTM_A$ and $LSTM_O$) are employed to aggregate and derive global representations separately for the actions and objects pathways, respectively, as presented in Eq. (\ref{eq:S_LSTM}). By employing bidirectional processing, the proposed approach integrates both past and future contextual information from the subtitle. It also integrates the most and least similar labels from the pretrained model features, facilitating a more nuanced and robust comprehension. 

\begin{equation}\label{eq:S_LSTM}
\begin{split}
\bar{S}_{A} = LSTM_A (\bar{S}_{A}; \theta_{LSTM_A}), \hspace{.5cm} 
\bar{S}_{O} = LSTM_O (\bar{S}_{O}; \theta_{LSTM_O}) \\
\bar{D}_{A} = LSTM_A (\bar{D}_{A}; \theta_{LSTM_A}), \hspace{.5cm}
\bar{D}_{O} = LSTM_O (\bar{D}_{O}; \theta_{LSTM_O}) \\
\bar{T}_{A} = LSTM_A (\bar{T}_{A}; \theta_{LSTM_A}), \hspace{.5cm}
\bar{T}_{O} = LSTM_O (\bar{T}_{O}; \theta_{LSTM_O}) \\
\end{split}
\end{equation}


\begin{figure}[tb]
  \centering
  \includegraphics[height=4.5cm]{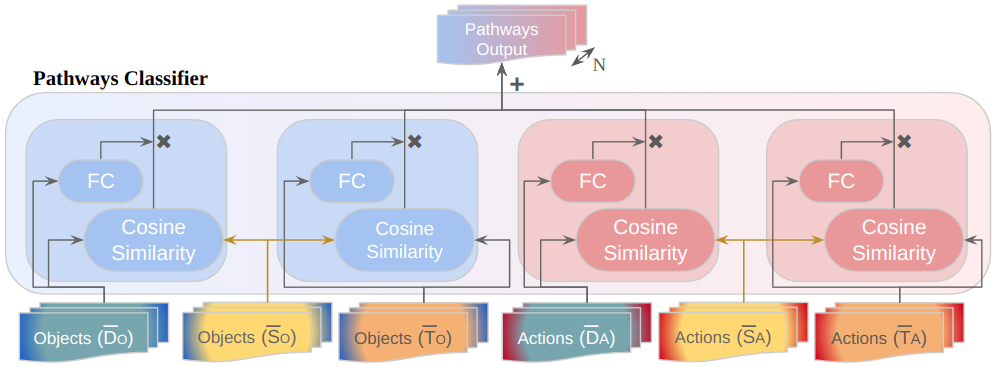}
  \caption{The pathways classifier of the proposed approach. 
  }
  \label{fig:pathways_classifier}
\end{figure}

\subsection{Actions and Objects Pathways Classifier}

The pathways classifier calculates the similarities between the clip-related and the answer-related features for each pathway to deduce the correct answer, as depicted in Fig. \ref{fig:pathways_classifier}. It assigns distinct attention weights to each cosine similarity output, where the weights are derived from a fully connected layer ($FC$), as presented in Eq. (\ref{eq:cosine_similarity}). The notions $\alpha_{\bar{D}_{A}}$, $\alpha_{\bar{D}_{O}}$, $\alpha_{\bar{T}_{A}}$, and $\alpha_{\bar{T}_{O}}$ are the attention weights of the action features and the object features of the audio modality and the text modality, respectively.
Finally, it sums up all the weighted similarity outputs obtained, as presented in Eq. (\ref{eq:sum_similarities}), where $P$ is the pathways output. 

\begin{equation}\label{eq:cosine_similarity}
\begin{split}
\bar{D}_{A} = CosineSimilarity (\bar{D}_{A}, \bar{S}_{A}) \times \alpha_{\bar{D}_{A}}, \hspace{.5cm} \alpha_{\bar{D}_{A}} = FC(\bar{D}_{A}; \theta_{FC})\\
\bar{D}_{O} = CosineSimilarity (\bar{D}_{O}, \bar{S}_{O}) \times \alpha_{\bar{D}_{O}}, \hspace{.5cm} \alpha_{\bar{D}_{O}} = FC(\bar{D}_{O}; \theta_{FC})\\
\bar{T}_{A} = CosineSimilarity (\bar{T}_{A}, \bar{S}_{A}) \times \alpha_{\bar{T}_{A}}, \hspace{.5cm} \alpha_{\bar{T}_{A}} = FC(\bar{T}_{A}; \theta_{FC})\\
\bar{T}_{O} = CosineSimilarity (\bar{T}_{O}, \bar{S}_{O}) \times \alpha_{\bar{T}_{O}}, \hspace{.5cm} \alpha_{\bar{T}_{O}} = FC(\bar{T}_{O}; \theta_{FC})\\
\end{split}
\end{equation}

\begin{equation}\label{eq:sum_similarities}
P = \bar{D}_{A} + \bar{D}_{O} + \bar{T}_{A} + \bar{T}_{O}\\
\end{equation}

\subsubsection{Text Features Classifier} The proposed AOPath also utilizes the raw pretrained backbone text features $T$ by employing a linear layer classifier $FC_T$, as illustrated in Fig. \ref{fig:framework} and presented in Eq. (\ref{eq:audio_text_classifier}). 
As shown in Eq. (\ref{eq:final_output}), the final multi-choice answer probability ($c_n$) is then derived through the summation of the outputs from the individual pathways and text classifiers.

\begin{equation}\label{eq:audio_text_classifier}
T = FC_T(T; \theta_{FC_T})
\end{equation}

\begin{equation}\label{eq:final_output}
\begin{split}
c_n =  T + P\\
\end{split}
\end{equation} 



\section{Experiments}

We validate the proposed approach on the TVQA dataset \cite{lei2019tvqa} to assess its generalization across various storytelling styles, thematic elements, character names, vocabulary, and narrative structures unique to each genre. The TVQA dataset is derived from six well-known TV series including three diverse genres: medical dramas, sitcoms, and crime shows. It encompasses 152.5K human-generated question-answer pairs and is characterized by its extensive and authentic nature, consisting of 21,793 video clips. To evaluate the robustness of the proposed approach to out-of-domain datasets, we partition the dataset into smaller subsets based on the distinct genres of the TV shows. These subsets, referred to as the medical subset, sitcoms subset, and crime subset, are denoted by the capital letters as $M$, $S$, and $C$ in the experiments, respectively. 
This deliberate segmentation facilitates a comprehensive evaluation of the VideoQA classifier's ability to generalize across different storytelling styles, thematic elements, and narrative structures inherent to each genre when trained on a specific genre and validated on another. Moreover, we further partition the genre-based subsets into smaller subsets, each containing videos from a single TV series and has approximately the same number of question-answer pairs. The downsized genre subsets, denoted by the small letters $m$, $s$, and $c$ in the experiments, include video clips exclusively from the TV series “House,” “Big Bang Theory,” and “Castle,” respectively. The three downsized genre subsets comprises 32,345, 29,384, and 32,886 question-answer pairs, respectively. Furthermore, the proposed classifier adapts the source genre (source domain) to the target genre (target domain) by utilizing parameters learned from the source genre. Specifically, the classifier is trained on the $X$ genre (source genre) and then tested on the $Y$ genre (target domain) without an explicit training on the $Y$ genre, relying solely on the weights learned from the $X$ genre. This approach demonstrates the classifier's ability to generalize to the $Y$ genre without direct training. In the experiments, the notation $X\rightarrow Y$ indicates training on the $X$ genre subset and validating the model on the $Y$ genre subset. The proposed approach is validated using three training datasets and tested across more than eight distinct training and validation settings.

\subsection{Hyperparameters and Experimental Setup}

The initial audio ($D$) and text ($T$) features are extracted from the base version of the large pretrained model MERLOT Reserve \cite{Zellers_2022_CVPR}, each of which has a size of 768. For the embedding of extracted verbs and nouns from the subtitles, the pretrained BPE embedding table (referred to as the Text Embedder in Fig. \ref{fig:framework} and Fig. \ref{fig:AOExtractor}) of the large pretrained model MERLOT Reserve is utilized, with an embedding size of 768. Similarly, the actions and objects dictionaries are embedded using the pretrained BPE embedding table. The actions dictionary comprises 1000 most common verbs used in English, sourced from the internet, while the objects dictionary consists of 1,374 object classes obtained from the Visual Genome dataset \cite{krishna2016visual}. We propose two versions of AOPath with different sizes: AOPath base version ($AOPath_B$) and AOPath small version ($AOPath_S$). They are both trained for 5 epochs, with a batch size of 32 and implemented using PyTorch with its default initializations. They are optimized by the Adam optimizer with a learning rate of 0.0003 \cite{kingma2017adam}. In $AOPath_B$ and $AOPath_S$, the linear layers $FC_{DA}$, $FC_{DO}$, $FC_A$ and $FC_O$ transform the pathway feature size from 768 to 256 and 8, while the modules $LSTM_A$ and $LSTM_O$ have a hidden size of 128 and 4, respectively. The total number of trainable weights in the proposed task-specific classifiers $AOPath_B$ and $AOPath_S$ are 1,579,010 and 26,282, respectively.

\subsection{Ablation study}

In the first raw of the first column of Table \ref{tab:ablation_study}, we evaluate the impact of varying the number of utilized action and object labels (denoted by $K$ in Fig. \ref{fig:AOExtractor}) on the overall model performance. The results indicate that utilizing the 15 most similar actions and objects leads to optimal performance across all settings. 
This suggests that employing the 15 most similar actions and objects results in a higher level of robustness and understanding. In the first raw of the second column of Table \ref{tab:ablation_study}, we assess the impact of the size of the pathway features obtained after the linear layers. In the $m\rightarrow  s$ setting, it is observed that 
both sizes 128 and 256 have the same average accuracy of $(63.42+63.44+66.99)/3=64.61$ and $(63.67+63.37+66.81)/3=64.61$ across all settings, respectively, suggesting a comparable performance.

\begin{table}
\begin{center}
\begin{tabular}{cccccccc}
\hline
 & $c\rightarrow m$ & $m\rightarrow s$ & $s\rightarrow c$ & & $c\rightarrow m$ & $m\rightarrow s$ & $s\rightarrow c$\\
 \hline
 \hline
No. Labels: 2 & 63.64 &  \textbf{63.37}& 66.75&Feature Size: 128& 63.42 & \textbf{63.44} & \textbf{66.99} \\
No. Labels: 15 & \textbf{63.67} & \textbf{63.37}& \textbf{66.81} & Feature Size: 256& \textbf{63.67} & 63.37& 66.81\\
No. Labels: 30 &  63.45& 63.08  & 66.72& Feature Size: 384& 63.23 & 63.08 & 66.60\\
\hline
\hline

Actions Pathway& 62.96 & 63.31 & 66.66 &Audio \& Pathways & 62.06 & 62.81&66.48\\
Objects Pathway & 63.14 & \textbf{63.41} & 66.75 & Text \& Pathways & \textbf{63.67} & \textbf{63.37}& 66.81 \\
Actions \& Objects  & \textbf{63.67} & 63.37& \textbf{66.81} &A \& T \& P & 62.83 & 62.98& \textbf{66.99}\\
\hline
\hline
Attention  & \textbf{63.67} & \textbf{63.37} & \textbf{66.81} & Text Output & 61.38 & 61.72 & 65.24\\
No-Attention & 62.55& 62.35 & 66.14 &Pathways Output  & \textbf{62.89} & \textbf{62.74} & \textbf{65.90}\\

\hline

\end{tabular}
\end{center}
\caption{Ablation study of the proposed $AOPath_B$  on the small genre subsets of the TVQA validation dataset. Each column represents training on a certain genre subset and then validating on another genre.}
\label{tab:ablation_study}
\end{table}

In the second row of the first column of Table \ref{tab:ablation_study}, we investigate the independent and joint utilization of the actions and objects pathways. 
The experimental results reveal that leveraging both pathways concurrently yields the most favorable results across most settings, thus achieving the highest level of robustness. However, In the $m\rightarrow  s$ setting, the independent utilization of the objects pathway achieves an accuracy of 63.41\%, outperforming the utilization of both pathways by a small margin of ($63.41 - 63.37=0.04$). This out-performance can be attributed to the specific characteristics of the medical genre, which often necessitates a deeper understanding of multiple objects rather than just actions. In the second row of the second column of Table \ref{tab:ablation_study}, we explore the joint and independent utilization of the Audio Output ($D$), the Text Output ($T$), and the Pathways Output ($P$). The text classifier and the Pathways classifier are illustrated in Fig. \ref{fig:framework}. The audio classifier, similar to the text classifier, processes the audio features ($D$). The results indicate that integrating the text output with the pathways output outperformed the independent incorporation of the audio and pathways output. Moreover, it outperformed the incorporation of all three outputs together in the remaining settings.

In addition, we explore the impact of the attention weights utilized in the pathways classifier. The third row of the first column reveals that removal of the attention weights lead to a degradation in performance across all different training and validation settings by significant margins of 1.12\%, 1.02\%, and 0.67\% in the $c\rightarrow  m$, $m\rightarrow  s$, and $s\rightarrow  c$ settings, respectively. 
We also conduct a thorough analysis in the third row of the second column of Table \ref{tab:ablation_study}, evaluating the individual performance of the Text Classifier, and the Pathways classifier. The results reveal that the proposed Pathways Classifier exhibits the highest accuracy in all training and validation settings. 

\begin{table}
\begin{center}
\begin{tabular}{cccccccccc}
\hline
 & $c\rightarrow m$ & $m\rightarrow s$ & $s\rightarrow c$& $c\rightarrow s$ & $m\rightarrow c$ & $s\rightarrow m$ & $c\rightarrow c$ & $m\rightarrow m$ & $s\rightarrow s$\\
\hline
Res. + $ATClassifier$ &60.51 & 61.12 & 65.18&  61.52& 66.14 & 60.17  & 66.41 & 62.18 & 63.51\\
Res. + $AOPath_S$ & 63.39 & \textbf{63.37} & 66.45& \textbf{63.90}& 67.99 & 61.22& 68.92 & 64.38 & 65.63\\
Res. + $AOPath_B$ & \textbf{63.67} & \textbf{63.37} & \textbf{66.81}& 63.61& \textbf{68.26} & \textbf{61.47}& \textbf{69.04} & \textbf{64.94} & \textbf{65.69}\\

\hline

\end{tabular}
\end{center}
\caption{Comparisons of different task-specific classifiers across all various training and validation settings on the small genre subsets of the TVQA dataset. The abbreviation "Res." stands for "Reserve".}
\label{tab:heads_comparisons}
\end{table}

\subsection{Comparisons of Different Approaches}

In Table \ref{tab:heads_comparisons}, we validate the proposed $AOPath_B$ and $AOPath_S$ on the genre-based subsets in comparison with a conventional linear layer classifier. The notation “$ATClassifier$” denotes the exclusive utilization of audio ($D$) and text features ($T$) from the large pretrained model, which are then classified through a simple linear layer head. 
The proposed $AOPath_B$ consistently outperforms the conventional linear layer classifier ($ATClassifier$) in all out-of-domain and in-domain settings with substantial margins 
It also demonstrates 5\% and 4\% superior performance over $ATClassifier$ in the $c\rightarrow  m$ and $m\rightarrow m$ settings, respectively. Moreover, $AOPath_B$ and $AOPath_S$ exhibit comparable performance across most training and validation settings.

Similarly, in Table \ref{tab:heads_comparisons_large_data}, we compare the previously introduced classifiers on the larger versions of the genre subsets, where the number of question and answer pairs varies greatly in each genre. The $No-Paths$ Classifier uses the same structure of LSTM and fully connected (FC) modules as the AOPath but does not incorporate the action and object pathways mechanism.. 
$Multi-Stream$ refers to a multi-stream trainable neural network \cite{lei2019tvqa} that utilizes features from subtitle, video, and visually present objects in the video clip. The video features are extracted from ResNet101 \cite{He_2016_CVPR}, trained on ImageNet \cite{5206848}, while visually present objects are extracted from Faster R-CNN \cite{NIPS2015_14bfa6bb}, trained on Visual Genome \cite{krishna2016visual}. The text embeddings of the question, multiple choice answers, subtitles, and visually present objects are obtained from the GloVe module \cite{pennington2014glove}. Moreover, $Multi-Stream$ has 14 million trainable parameters, making it significantly larger than $AOPath_B$ and $AOPath_S$, with 12.5 and 14 million more trainable parameters, respectively.

Similar to $AOPath_B$, the proposed $AOPath_S$ demonstrates superior performance compared to $Multi-Stream$ across five out-of-domain datasets, achieving high margins of improvement in the settings of $C\rightarrow  M$, $M\rightarrow  S$, $S\rightarrow  C$, $C\rightarrow  S$, and $M\rightarrow  C$
It also demonstrates 3\% superior performance over $Multi-Stream$ in the $M\rightarrow  S$, $C\rightarrow S$ and $M\rightarrow C$ settings. Conversely, $Multi-Stream$ outperforms $AOPath_S$ in the $S\rightarrow  M$ setting, albeit with a smaller margin of $63.71-63.43=0.28$, despite having 14 million more trainable parameters. This discrepancy suggests that $AOPath_S$ neatly reasons over the pretrained features by extracting domain-agnostic action and object features, thereby mitigating exposure to out-of-domain features. In in-domain settings, $Multi-Stream$ outperforms $AOPath_S$ by only 0.1\% and 1\% with margins of $0.09\%$ and $0.73\%$ in the $C\rightarrow C$ and $M\rightarrow M$ settings, respectively, which suggests the competitiveness of $AOPath_S$. It also indicates that $AOPath_S$ is acquiring knowledge rather than exploiting spurious correlations within the dataset, resulting in competitive performance in both in-domain and out-of-domain scenarios. 

\begin{table}
\begin{center}
\begin{tabular}{ccccccccc}
\hline
 & $C\rightarrow M$ & $M\rightarrow S$ & $S\rightarrow C$ & $C\rightarrow S$ & $M\rightarrow C$ & $S\rightarrow M$  & $M\rightarrow M$ & $S\rightarrow S$\\
 \hline
 
$ATClassifier$ & 60.65 & 61.33 & 66.57  & 60.90  & 66.29 & 61.79& 62.95& 62.92 \\
$No-Paths$& 61.21& 61.28 & 66.84& 61.13& 66.60& 62.12 & 63.53 & 62.92 \\
$Multi-Stream$& 62.47 &  60.81& 66.87& 60.94& 66.41 & \textbf{63.71} & \textbf{65.90} & \textbf{66.78}\\

$AOPath_B$ & 63.65 & 62.78 & 68.32& 62.74& 68.02& 63.30 & 64.69& 64.81 \\

$AOPath_{S}$ & \textbf{63.68} & \textbf{62.88} & \textbf{68.59}& \textbf{62.83} & \textbf{68.95}& 63.43& 65.17& 65.16\\

\hline
\end{tabular}
\end{center}
\caption{Comparisons of different task-specific classifiers across all various training and validation settings on the large genre subsets of the TVQA dataset.}

\label{tab:heads_comparisons_large_data}
\end{table}

\begin{figure}[ht]
\begin{center}

\includegraphics[width=1.0\linewidth]{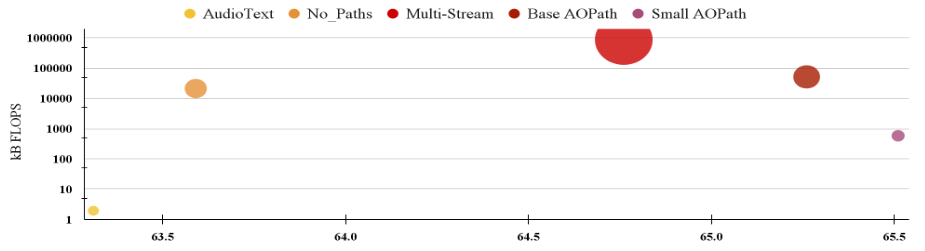}
\end{center}
 \caption{
 Comparison of different classifiers with respect to average accuracy, floating-point operations (FLOPs), and the number of trainable weights}
\label{fig:flops}
\end{figure}

\textbf{Comparisons in Terms of Complexity} As shown in Fig. \ref{fig:flops}, the $AOPath_S$ has the highest average accuracy of 65.51\% across all validation settings, including out-of-domain and in-domain, in the large genre subsets of the TVQA dataset. Moreover, it has significantly a lower number of trainable parameters of 26,282, as indicated by the size of the circle, compared to $Multi-Stream$, which has 14,133,303 trainable parameters. On the other hand, $ATClassifier$ has the lowest number of trainable parameters, which is 769, but the lowest average accuracy. In terms of floating-point operations (FLOPs), $AOPath_S$ requires only 582kB FLOPs to generalize to out-of-domain datasets efficiently, while $Multi-Stream$ struggles to generalize to out-of-domain, requiring 871,730kB FLOPs. 

\subsection{Qualitative results}

\begin{figure}[ht]
\begin{center}

\includegraphics[width=1\linewidth]{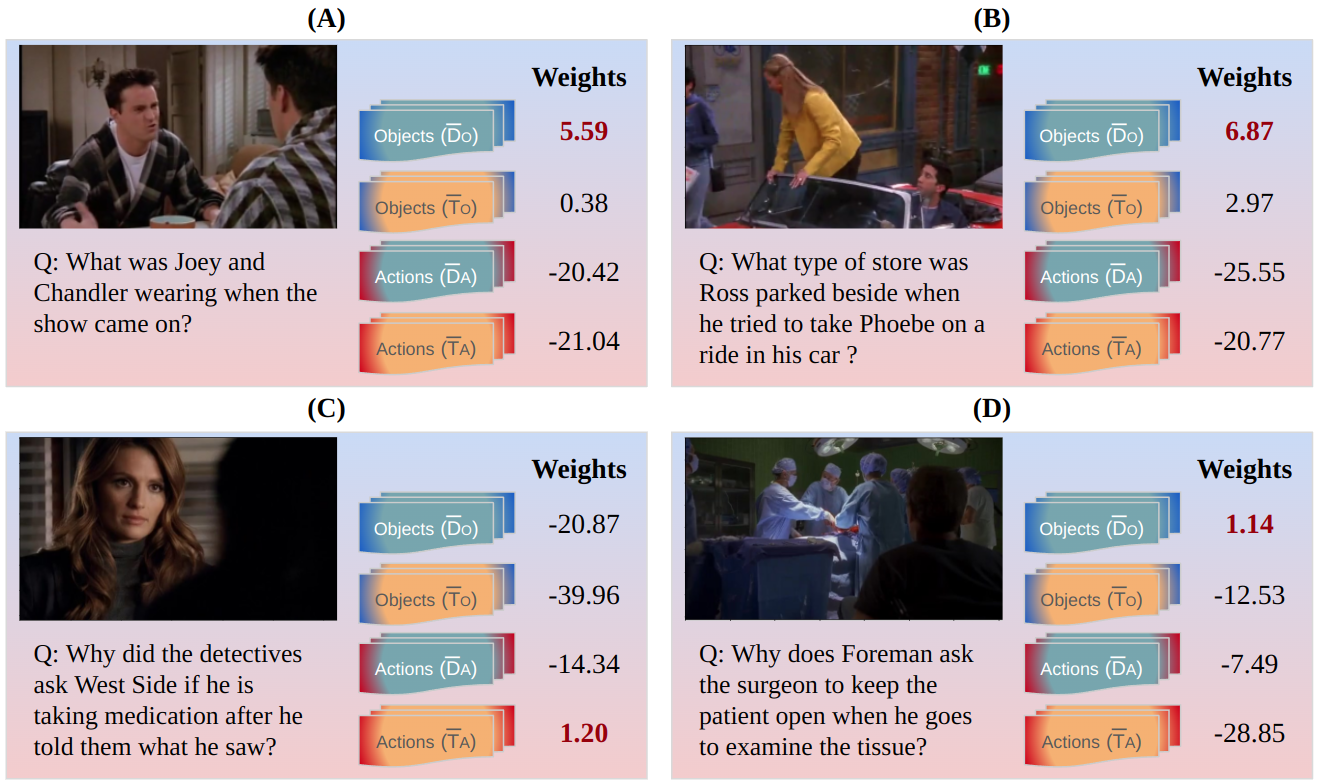}
\end{center}
 \caption{
 Qualitative results of attention weights in the pathways classifier. The green-colored boxes display the pathway weights for audio features, while the orange-colored boxes depict the pathway weights for text features. Boxes outlined in blue represent the weights of the object pathways, and those outlined in red represent the weights of the action pathways.}
\label{fig:q_0}
\end{figure}

We conduct qualitative analysis of the attention weights utilized in the pathways classifier depicted in Fig. \ref{fig:pathways_classifier}. As illustrated in (A) of Fig. \ref{fig:q_0}, the pathways classifier assigns varying attention weights to each pathway. Notably, it assigns the highest attention weight of 5.59 to the object pathway of the audio modality. Considering the query “What were Joey and Chandler wearing when the show came on?”, the question requires more reasoning over objects than actions in late stages to identify the correct answer. Similarly, as illustrated in (B) of Fig. \ref{fig:q_0}, the pathways classifier assigns the highest attention weights of 6.87 and 2.97 to the objects pathways. Considering the question “What type of store was Ross parked beside when he tried to take Phoebe on a ride in his car?”, it primarily focuses on identifying an object or a physical attribute rather than an action or motion.

However, in (C) of Fig. \ref{fig:q_0}, the pathways classifier assigns the highest weights of 1.20 and -14.34 to the actions pathway. Such “why” questions typically seek explanations or reasons, indicating a need to focus more on actions rather than objects. Similarly, in (D) of Fig. \ref{fig:q_0}, the pathways classifier assigns the highest attention weights of 1.14 and -7.49 to the objects pathway followed by the actions pathway of audio. Such query, seeking explanations or reasons, necessitate a deeper level of reasoning, particularly regarding actions.

\section{Conclusion}

In this paper, we introduced the Actions and Objects Pathways (AOPath), a novel task-specific classifier in the domain of VideoQA designed for robust generalization to out-of-domain datasets. Inspired by the human brain, it achieved computationally inexpensive robustness by dissociating features into action and object features and by creating separate pathways for each. It also employed the AOExtractor module, which introduces no trainable weights and transforms out-of-domain into domain-agnostic features. We demonstrated the robustness of the proposed approach for VideoQA on the TVQA dataset. Furthermore, we demonstrated the efficiency of the actions and objects pathways design through an extensive ablation study and qualitative results. The proposed AOPath achieved superior results on out-of-domain genre subsets compared to other methods, despite having significantly fewer trainable weights.

\subsubsection{\ackname} This research was supported in part by the MSIT (Ministry of Science and ICT), Korea, under the ITRC (Information Technology Research Center) support program (IITP-2024-2020-0-01808) supervised by the IITP (Institute of Information \& Communications Technology Planning \& Evaluation), and in part by Basic Science Research Program through the National Research Foundation of Korea (NRF) funded by the Ministry of Education (2021R1A6A1A03043144).

\bibliographystyle{splncs04}
\bibliography{0411.bib}
\end{document}